\documentclass[twoside,leqno,twocolumn]{article}

\usepackage[letterpaper]{geometry}

\usepackage{siamproceedings}
\hypersetup{hidelinks}

\usepackage[normalem]{ulem} 
\ifpdf
  \DeclareGraphicsExtensions{.eps,.pdf,.png,.jpg}
\else
  \DeclareGraphicsExtensions{.eps}
\fi

\newsiamremark{remark}{Remark}
\newsiamremark{hypothesis}{Hypothesis}
\crefname{hypothesis}{Hypothesis}{Hypotheses}
\newsiamthm{claim}{Claim}

\usepackage{amssymb,amsfonts}
\usepackage{algorithmic}
\usepackage{booktabs}
\usepackage{subcaption}
\usepackage{color}
\usepackage{float}
\usepackage[T1]{fontenc}
\usepackage{enumitem}
\usepackage{soul}

\graphicspath{
  {./}
  {./results/}
  {./results/multidata/}
  {./OverclusterDRG_Paper/}
  {./OverclusteringRKC/}
  {./overclustering_me/}
}

\newcommand{\OPT}{\mathrm{OPT}}

\newcommand{\rkc}{R$k$C}

\newcommand{\N}{n}

\begin{document}

\newcommand\relatedversion{}

\title{\Large OverclusterDRG: a Deterministic, LP-Free Heuristic for Robust \texorpdfstring{$k$}{k}-Center\relatedversion}
\author{
  Rudra Bhardwaj\thanks{Department of Computer Science, University of Delhi (\email{rudramsc24@cs.du.ac.in}, \email{ngupta@cs.du.ac.in}).}
  \and Rajni Dabas\thanks{Department of Computer Science, Northwestern University; P.G.D.A.V. College, University of Delhi (\email{rajni.dabas@pgdav.du.ac.in}).}
  \and Neelima Gupta\footnotemark[1]
}

\date{}

\maketitle

\fancyfoot[R]{\scriptsize{Copyright \textcopyright\ 20XX by SIAM\\
Unauthorized reproduction of this article is prohibited}}

\begin{abstract} 
The $k$-center problem is a fundamental formulation for metric clustering. However, its min-max objective is highly sensitive to outliers, motivating the Robust $k$-Center (\rkc) formulation which permits excluding up to $z$ points as outliers. 

Existing algorithms for \rkc~suffer from scalability issues: LP rounding relies on fractional relaxations that scale poorly \cite{harris2017lottery,chakrabarty2016non}, known combinatorial approximation algorithms are computationally expensive due to radius guessing \cite{charikar2001}, and scalable randomized heuristics require exponential trial multipliers to guarantee success \cite{ding2019}. 

We present OverclusterDRG, a deterministic, LP-free heuristic for \rkc. The algorithm leverages the theoretical insight of farthest-point oversampling - adding exactly $z$ extra candidates to center set to absorb outliers - followed by a forward-greedy initialization and a $1$-swap local search. Evaluations across real-world datasets show that OverclusterDRG attains a mean approximation ratio of 1.147 against fractional LP lower bounds on the UCI Adult dataset, outperforming state-of-the-art baselines in both clustering quality and speed. By avoiding subset-guessing and exponential trial multipliers, it scales smoothly to $n=10^5$ points where LP solvers and theoretical algorithms become intractable.
\end{abstract}

\section{Introduction}

The $k$-center problem is a foundational framework in metric clustering and facility location. Given a dataset with $n$ points and a budget of $k$ centers, the objective is to place the centers such that the maximum distance from any data point to its nearest center is minimized. Because it optimizes a worst-case distance, $k$-center is the standard formulation for applications requiring strict service guarantees, such as emergency facility placement, sensor coverage, and bounding network latency. While the exact problem is NP-hard, Gonzalez \cite{gonzalez1985} introduced an elegant $O(nk)$-time greedy algorithm that achieves a tight 2-factor approximation. The algorithm operates via a farthest-point traversal: starting with an arbitrary point, it iteratively selects the point farthest from the current set of centers until $k$ centers are picked.

However, in the rigid min-max objective of the classical $k$-center problem, a single distant outlier, measurement error, or anomalous data entry can dictate the objective. It can drag center away from dense, meaningful clusters merely to cover the isolated point. To handle this, the Robust $k$-Center problem (\rkc) introduces an outlier budget $z \ge 0$. The algorithm is allowed to discard up to $z$ points from the dataset, and the objective is evaluated only on the remaining $(n - z)$ non-outlier points. While this robust formulation protects the clustering from noise, it introduces a significant computational burden: identifying the optimal subset of outliers superimposes a combinatorial selection problem on top of an already NP-hard clustering task.

While \rkc~elegantly models real-world noise, algorithms for solving it reveal a severe disconnect between theoretical worst-case guarantees and practical algorithm engineering. A natural practitioner baseline is to run the Gonzalez traversal for $k$ steps and evaluate the radius after discarding the $z$ farthest points. Unfortunately, all theoretical guarantees vanish under this approach: a single outlier can irreparably hijack a center selection, leaving true clusters entirely unrepresented. Consequently, the theoretical literature has developed approximation algorithms with strict formal guarantees, but these algorithms present computational bottlenecks at scale. We sketch them as they form our baselines.
\begin{itemize}
     \item \textbf{Combinatorial Approximation:} Charikar et al.~\cite{charikar2001} introduced the first constant-factor approximation for robust $k$-center, yielding a 3-factor approximation with running time of $O(kn^2\log n)$. While theoretically significant, their algorithm relies on optimal radius-guessing, constructing a graph of intersecting disks, and greedily selecting expanded clusters. Our empirical profiling reveals that this method scales as $O(n^{2.12})$  even in practice, requiring hundreds of seconds on moderately large datasets.
     
    \item \textbf{LP rounding schemes:} The approximation factor for \rkc~was later improved to a tight 2 using linear programming (LP) rounding by Chakrabarty et al. \cite{chakrabarty2016non}; Harris et al. \cite{harris2017lottery}. While these LP-based approaches yield optimal approximation guarantees, they suffer from scalability challenge. The underlying fractional LP relaxation requires modeling the assignment between every client and every potential center, resulting in $O(n^2)$ variables and $O(n^2)$ constraints. Theoretically, the current state-of-the-art time complexity for solving a general linear program is $\tilde{O}(V^\omega)$, where $V$ is the number of variables and $\omega \approx 2.37$ is the matrix multiplication exponent \cite{cohen2021solving}. Consequently, evaluating one instance of \rkc~LP demands an $\tilde{O}(n^{4.74})$ time theoretically. Our evaluations show that in practice also solving an LP requires up to 94 hours for a single configuration at $\N=20,000$.
    
    \item \textbf{Randomized sampling:} To bypass these computational bottlenecks, Ding et al. \cite{ding2019} proposed a randomized greedy strategy that achieves a $2$-approximation with a constant probability per trial. While a single execution takes just $\mathcal{O}(kn)$ time, achieving their formal approximation guarantee requires a sequence of successful probabilistic choices. To strictly respect an exact outlier budget of $z$, the algorithm must sample from exactly the $z+1$ farthest candidates (equivalent to setting their parameter $\epsilon=1/z$). Under this strict parameterization, guaranteeing a 99\% probability of success formally requires $T = \lceil \ln(100)(\frac{n}{n-z}) (z+1)^{k-1} \rceil$ independent trials. This exponential dependence on $k$ quickly inflates the overall complexity to $O(nk2^k)$. While this $\mathcal{O}(2^k)$ scaling is manageable for small cluster counts (Ding et al. notably restrict their exact evaluations to $k \le 20$), it becomes computationally intractable for larger configurations. For example, on a dataset with a 20\% outlier rate and $k=10$ clusters, Ding's algorithm theoretically requires roughly 2,900 passes, and $k=50$ demands an intractable $\sim 10^{15}$ passes. Consequently, as the number of clusters grows, practitioners are forced to truncate the randomized search, abandoning the formal theoretical guarantee for speed.
    
   \item  \textbf{Natural Heuristics:} Given that algorithms with provable guarantees are not practical, a natural fast heuristic is a simple adaptation of Gonzalez's algorithm: at each iteration, skip the $z$ most extreme points and instead selects the $(z+1)^{\text{th}}$ farthest point as the next center. Once all $k$ centers are placed, the algorithm assigns every point to its nearest center and discards the $z$ points with the largest distances as outliers. Computing $(z+1)^{\text{th}}$ farthest point takes $O(n \log z)$ time per iteration. This yields a efficient overall time complexity of $O(kn \log z)$. However, this strategy offers no formal approximation guarantees and yields poor quality on real-world datasets.

\end{itemize}

There remains a need for an algorithm that provides the empirical quality of combinatorial and LP-based methods, operates deterministically, and scales to large datasets with low computational overhead.

In this paper, we present \textbf{OverclusterDRG} - \textbf{Over}clustering
followed by a \textbf{D}eterministic \textbf{R}obust \textbf{G}reedy
selection; a deterministic, LP-free heuristic for \rkc. We leverage the theoretical insight of candidate oversampling - a standard data reduction technique in the parameterized complexity (FPT) literature - but replace its subset-guessing step with a heuristic pipeline. 
Our algorithm operates in three steps. First, it deterministically builds a candidate pool of $k+z$ points using Gonzalez farthest-point rule (Step 1). Outliers will naturally attract the farthest-point selection, but by providing exactly $z$ spare steps, the pool can absorb them and still retain a representative center for every true cluster. Second, it extracts $k$ centers from this pool using a forward-greedy initialization (Step 2). Third, it refines the selection using a 1-swap local search algorithm (Step 3).

Our primary contributions are:
\begin{enumerate}
    \item Empirical Quality: To baseline heuristic performance, we computed LP lower bounds across 102 distinct configurations on three primary real-world datasets. On the UCI Adult dataset, OverclusterDRG achieves a mean LP approximation ratio of 1.147, outperforming the achieved ratio of both Charikar's (1.173) and Ding's (1.235) implementations. On a configuration-by-configuration basis, OverclusterDRG consistently dominates, attaining a tighter radius than Charikar in 33 of 36 cases, and outperforming Ding in 35 of 36 cases. This superior clustering quality reliably replicates across our secondary datasets, yielding mean approximation ratios of 1.155 on Diabetes and 1.171 on Covertype.
    
    \item  Computational Scalability: OverclusterDRG achieves these tighter bounds while operating at a fraction of the computational cost of existing methods. On the Adult dataset, our algorithm completes with a median runtime of just 0.36 seconds, yielding a $10\times$ speedup over Charikar (3.60s) and a $6\times$ speedup over Ding (2.22s). By utilizing memory-efficient distance matrix caching, OverclusterDRG yields an empirical time complexity that scales at roughly $O(\N^{1.85})$. At large scales ($\N = 99,000$), where LPs and Charikar's algorithm become computationally intractable, OverclusterDRG remains highly efficient, completing execution in 1,100 seconds (more than three times faster than Ding's randomized baseline of 3,800 seconds).
    
    \item Validating impact of Step 3: To validate the internal mechanics of our heuristic, we evaluated our 1-swap local search against an exhaustive combinatorial evaluation on 400 small subsampled instances. OverclusterDRG perfectly recovered the exact combinatorial pool optimum in 97.2\% of instances. In the rare cases where it settled into a local minimum, the maximum observed optimality gap was only 4.82\%. Furthermore, the experiments confirm that this 1-swap polish provides a statistically significant quality improvement ($p < 10^{-8}$) over the greedy initialization alone in 36 of 36 configurations, fully justifying our step 3.
    
    \item Validation of the Overcount in Step 1: We conduct an empirical study on the size of the oversampled candidate pool, revealing a sharp phase transition in clustering quality. Across 72 configurations, we demonstrate that the clustering radius degrades severely when the pool is starved ($< z$), but flattens into a tight performance exactly as it reaches and surpasses $z$.
\end{enumerate}

\subsection{Related Work}
\label{sec:related}

Charikar et al.~\cite{charikar2001} introduced the first constant-factor ($3$) approximation for robust $k$-center based on a greedy approach, which was later improved to a tight factor of $2$ using LP rounding \cite{chakrabarty2016non,harris2017lottery}. In 2019, Ding et al.~\cite{ding2019} showed that a randomised greedy strategy succeeds with a constant probability per trial, which can be amplified to a $99\%$ success rate using $T \approx \lceil \ln(100) (\frac{n}{n-z}) 2^{k-1} \rceil$ independent randomized trials. On the parameterized complexity side, Goyal and Jaiswal~\cite{goyal2021} gave tight approximations where the running time is fixed-parameter tractable (FPT) in $k$ for constrained $k$-center variants, which include \rkc.

Other clustering objectives like $k$-median (minimizing the sum of distances), $k$-means (minimizing the sum of squared distances), and facility location (incorporating facility opening costs instead of a hard bound $k$, minimizing the sum of distances plus facility costs) have also been extensively studied in the robust framework. 

The robust $k$-median problem (R$k$M) was first studied by Charikar et al.~\cite{charikar2001}, who proposed a bi-criteria approximation algorithm achieving a cost approximation factor of $4(1+\frac{1}{\epsilon})$ while allowing a $(1+\epsilon)$ violation in the number of outliers. The first true constant-factor approximation for $k$MO was obtained by Chen~\cite{chen_constant_2008}. However, the approximation factor in Chen's result, while constant, is relatively large and unspecified. Krishnaswamy et al.~\cite{krishnaswamy_constant_2017} significantly improved upon this by giving a $7.081$-approximation via an iterative LP rounding framework. This result was further refined by Gupta et al.~\cite{Gupta_kmo_2021}, who improved the approximation factor to $6.994 + \epsilon$. Furthermore, Friggstad et al.~\cite{Friggstad_LS_2019} analyzed local search for this problem, providing a $(3+\epsilon)$-factor approximation with a $(1+\epsilon)$-factor violation in the cardinality ($k$).

For robust $k$-means, the theoretical trajectory closely mirrors that of robust $k$-median. Chen~\cite{chen_constant_2008} similarly provided the first strict constant-factor approximation. This was vastly improved by Krishnaswamy et al.~\cite{krishnaswamy_constant_2017}, whose iterative rounding framework achieved a $53.002$-approximation. The efficacy of local search for this sum-of-squares objective was also formalized by Friggstad et al.~\cite{Friggstad_LS_2019}, who proved that a standard multi-swap local search algorithm yields a constant-factor approximation for robust $k$-means, again relying on a $(1+\epsilon)$ violation in the number of clusters.

The concept of outliers in the facility location problem was also first introduced by Charikar et al.~\cite{charikar2001}, who developed a $3$-factor approximation algorithm. This approximation ratio was later improved to a factor of $2$ by Jain et al.~\cite{jain_greedy_2003}. 

\subsection{Organization of the paper}
The remainder of the paper is organized as follows. In Section~\ref{sec:algorithm}, we formally define the Robust $k$-Center problem and detail the OverclusterDRG algorithm. In Section~\ref{sec:experiments}, we present our empirical evaluations. We conclude in Section~\ref{sec:conclusion}.

\section{OverclusterDRG algorithm for \texorpdfstring{\rkc}{RkC}}
\label{sec:algorithm}
In this section, we present our algorithm for the \rkc~problem, formally defined as follows. Let $X \subset \mathbb{R}^d$ be a finite dataset of $n$ points equipped with a distance metric $d$. Given an integer cluster budget $k\ge 1$ and an outlier budget $z\ge 0$, the objective is to find a set of centers $C\subseteq X$ with $\vert{}C\vert{}=k$ and a set of outliers $S\subseteq X$ with $\vert{}S\vert{}\le z$ minimizing: $$\max_{x \in X \setminus S} d(x, C)$$ where $d(x, C) = \min_{c \in C} d(x, c)$.

Equivalently, for any set of candidate centers $C$, let $r_z(C, X)$ denote the robust radius, defined as the $(z+1)^{\text{th}}$ largest distance from a point in $X$ to its nearest center in $C$. The objective is to find $C \subseteq X$ with $|C|=k$ that minimizes $r_z(C, X)$.

The algorithm is driven by the insight that a standard Gonzalez farthest-point traversal inherently wastes steps on covering isolated outliers. By adding the extra steps, the candidate pool can absorb the outliers while retaining representative centers for true clusters. The algorithm operates in three steps: ($i$) building an overcomplete candidate pool, $(ii)$ initializing a solution greedily, and $(iii)$ polishing it with a 1-swap local search. We name the method OverclusterDRG (Overclustering followed by a Deterministic Robust Greedy selection). The unified pseudocode is provided in Algorithm \ref{alg:overcluster}. 

Following are the details of the three steps.

\textbf{Step 1: Overclustering (OC).} We run the Gonzalez farthest-point rule for $k+z$ steps. To ensure the algorithm is strictly deterministic, the first point $q_1$ is selected as the data point nearest to the geometric centroid of $X$ (i.e., $\bar{x} = \frac{1}{n}\sum_{x \in X} x$). Every subsequent point is the one farthest from the current pool: $q_{i+1} = \arg \max_{x \in X} \min_{j\le i} d(x,q_j)$. Ties in both argmins and argmaxes are broken lexicographically by point coordinate vectors in $\mathbb{R}^d$, guaranteeing an output that is independent of input sequence order. The resulting pool is $Q=\{q_1,\dots,q_{k+z}\}$. 

\textbf{Step 2: Forward greedy (Greedy).} We select an initial set of $k$ centers from $Q$. First, we precompute an $n\times(k+z)$ distance matrix mapping all $x \in X$ to the pool candidates $q \in Q$. Starting from the initial center set $C=\{q_1\}$, we iteratively evaluate all remaining pool candidates and add the point $q^*$ whose inclusion minimizes the robust radius $r_z(C \cup \{q^*\}, X)$. Ties are broken by selecting the candidate with the lowest pool index. This greedy addition is repeated until exactly $k$ centers are selected. The intermediate output is a strong heuristic on its own, which we report as {\em OC-Greedy}.

\textbf{Step 3: 1-swap local search (Local Search).} Finally, we refine the selection using a local search strictly limited to the pool $Q$. We apply a first-improvement strategy: for each active center $c \in C$ and each unused candidate $q \in Q\setminus C$, scanned in a fixed pool-index order, we evaluate the swap replacing $c$ with $q$. We accept the first swap that strictly decreases the robust radius by at least a numerical tolerance $\epsilon>0$, and immediately restart the scan. This process repeats until a swap yields no improvement.

\begin{algorithm}[ht]
\caption{OverclusterDRG($X, k, z$)}
\label{alg:overcluster}
\begin{algorithmic}[1]
\STATE \textbf{// Step 1: Pool Construction}
\STATE Compute the geometric centroid $\bar{x} = \frac{1}{n}\sum_{x \in X} x$
\STATE Let $q_1 \in X$ be the point nearest to $\bar{x}$ (ties broken lexicographically)
\STATE $Q \gets \{q_1\}$
\WHILE{$|Q| < k + z$}
    \STATE $q \gets \arg\max_{x \in X} \min_{p \in Q} d(x, p)$ 
    \STATE $Q \gets Q \cup \{q\}$
\ENDWHILE
\STATE \textbf{// Step 2: Greedy}
\STATE Precompute distance matrix $D[x][q] = d(x,q)$ for all $x \in X, q \in Q$
\STATE $C \gets \{q_1\}$
\WHILE{$|C| < k$}
    \STATE $q^* \gets \arg\min_{q \in Q \setminus C} r_z(C \cup \{q\}, X)$ 
    \STATE $C \gets C \cup \{q^*\}$
\ENDWHILE
\STATE \textbf{// Step 3: Local Search}
\REPEAT
    \STATE $\text{improved} \gets \text{False}$
    \FOR{\textbf{each} $c \in C$ \textbf{and} $q \in Q \setminus C$ }
        \STATE $C' \gets (C \setminus \{c\}) \cup \{q\}$
        \IF{$r_z(C', X) \le r_z(C, X) - \epsilon$}
            \STATE $C \gets C'$
            \STATE $\text{improved} \gets \text{True}$
            \STATE \textbf{break} 
        \ENDIF
    \ENDFOR
\UNTIL{\textbf{not} improved}
\RETURN $C$
\end{algorithmic}
\end{algorithm}

\textbf{Complexity.} By precomputing distances and restricting the search space to a constrained pool of size $k+z$, the runtime overhead is tightly bounded:
\begin{itemize}
    \item Step 1 maintains an array of the minimum distances from each of the $n$ points to the growing pool. When a new point $q$ is added, updating this array takes $\mathcal{O}(n)$ time. Finding the maximum over $X$ also takes $\mathcal{O}(n)$ time. We select exactly $k+z$ centers and therefore, constructing the entire pool requires $\mathcal{O}(n(k+z))$ time.

    \item Step 2 computes the initial dataset-to-pool distance matrix in $\mathcal{O}(n(k+z))$ time. In each of the $k-1$ subsequent selection steps, evaluating $r_z(C \cup \{q\}, X)$ for a candidate takes $\mathcal{O}(n)$ time using the cached matrix and a running minimum. Minimizing over all available pool candidates requires $\mathcal{O}(n(k+z))$ time per step. Selecting all $k$ centers therefore evaluates in a total of $\mathcal{O}(kn(k+z))$ time.

    \item Step 3 evaluates candidate swaps. There are $k$ active centers and $z$ unused candidates in the pool, resulting in up to $k \times z$ possible swaps per iteration of the \texttt{repeat} loop. Thus, a single iteration evaluating all pairs takes expected $\mathcal{O}(k^2 n z)$\footnote{Can be optimized by maintaining a standard caching data structure.} time. Because the objective strictly decreases by at least an absolute tolerance $\epsilon > 0$ per accepted swap, the maximum number of iterations is also strictly bounded.
\end{itemize}

Note that space complexity is governed by the $\mathcal{O}(n(k+z))$ storage required for the distance matrix cache.

\section{Experiments}
\label{sec:experiments}

In this section, we empirically evaluate OverclusterDRG against both theoretical algorithms and fast heuristics. We structure our evaluation into four modules: (1) Datasets, (2) Baselines and setup, (3) Results on clustering quality, scalability, and local search validity, and (4) Validation of the choice of overpicking by exactly $z$ candidates. All source code, logs, and charts are publicly available\footnote{\href{https://anonymous.4open.science/r/overclusterdrg-5122/}{https://anonymous.4open.science/r/overclusterdrg-5122/}}.

\subsection{Datasets}

We evaluate our algorithm on three standard real-world tabular datasets commonly used in the metric clustering literature (Table~\ref{tab:datasets}). For all datasets, categorical attributes are dropped, numerical features are $z$-score standardized, and distances are computed using the Euclidean metric.

\begin{table}[ht]
\caption{Datasets used in experiments. All numerical features are $z$-score standardized.}
\label{tab:datasets}
\centering
\resizebox{\columnwidth}{!}{%
\begin{tabular}{lrrl}
\toprule
Dataset & Rows Used & Features & Source \\
\midrule
UCI Adult & 32{,}561 & 6 & \cite{adult_dataset} \\
Diabetes Indicators & 99{,}492 & 8 & \cite{diabetes_dataset} \\
Covertype & 30{,}000 of 581{,}012 & 10 & \cite{covertype_dataset} \\
\bottomrule
\end{tabular}%
}
\end{table}

To systematically test algorithmic scalability across dataset sizes ($N$) while keeping the underlying data distributions mathematically consistent, we evaluate all algorithms on nested subsets of size $N$ (the datasets were globally shuffled using a fixed random seed). This subsampling is strictly necessary because evaluating the LP lower bound at full scale is computationally intractable, taking hours to days even for moderate $N$. 

For the UCI Adult and Diabetes datasets, the $N$-sized subset is generated deterministically by selecting the first $N$ rows of the standardized dataset. For Covertype, because the original dataset is excessively large, we first draw a fixed, uniform random sample of $30{,}000$ rows without replacement (using NumPy \texttt{default\_rng} with seed 42); the $N$-sized subsets are then constructed by selecting the first $N$ rows of this fixed sample. This strict nested-subset protocol ensures that for any given configuration of $(N, k, z/N)$, every evaluated algorithm - as well as the LP oracle - operates on the exact same identical data points, permitting perfectly paired comparisons.

Our primary evaluation grid (Adult) spans dataset sizes $N \in \{500, 1000, 2000, 5000, 10000, 20000\}$, cluster budgets $k \in \{10, 20, 50\}$, and outlier rates $z/N \in \{10\%, 20\%\}$, yielding 36 distinct configurations. For Diabetes, we use the identical grid but substitute $N=15{,}000$ for $N=2{,}000$ to test denser configurations (36 configurations). Covertype uses $N \in \{500, 1000, 5000, 10000, 15000\}$ across the same $k$ and $z/N$ variations (30 configurations).

\subsection{Baselines and Setup}

\paragraph{Hardware Configuration.} 
All experiments, including LP solutions, were executed sequentially on a single machine: NVIDIA RTX 5000 Ada (32 GB GPU VRAM) paired with 256 GB system RAM and an Intel Xeon W7-2475X processor.

\paragraph{LP Oracle Lower Bound.}
To accurately measure approximation quality, we benchmark all algorithms against the optimal fractional radius. For a fixed radius $r$, let $\mathcal{N}(j,r) = \{i \mid d(i,j) \le r\}$. We define variables $x_{ij} \in [0,1]$ (assignment of $j$ to center $i$) and $y_i \in [0,1]$ (center opened at $i$). We check the feasibility of the following standard relaxation:
\begin{align*}
\sum_{i \in \mathcal{N}(j,r)} x_{ij} &\le 1 \quad \forall j \\
x_{ij} &\le y_i \quad \forall i,j \\
\sum_{i} y_i & \le k \quad  \\
\sum_{j} \sum_{i \in \mathcal{N}(j,r)} x_{ij} & \ge N-z 
\end{align*}
A binary search over all pairwise distances yields the smallest feasible radius $R_{\mathrm{LP}} \le \OPT$. We solve the LP using Gurobi (tolerance $10^{-6}$). Computing this across all configurations takes up to 94 hours for $N=20{,}000$, but guarantees our reported approximation ratios ($R_{\mathrm{alg}} / R_{\mathrm{LP}}$) form upper bounds on the true ratio against $\OPT$.

A binary search over all pairwise distances yields the smallest feasible radius $R_{LP} \le OPT$. We evaluate LP feasibility at each step using Gurobi (numerical tolerance $10^{-6}$). The computational cost explodes rapidly: requiring $\approx 417$ seconds at $N=2{,}000$, $>36$ hours at $N=5{,}000$, and up to $94$ hours at $N=20{,}000$ (an empirical scaling of $\mathcal{O}(N^{3.40})$). Computing this across all configurations provides highly reliable numerical upper bounds on the true approximation ratios ($R_{\mathrm{alg}} / R_{\mathrm{LP}}$), while practically demonstrating why fractional LP-based solutions are unusable at scale.

\paragraph{\textbf{Algorithms Compared.}}
We compare OverclusterDRG (the full algorithm) and its greedy initialization {\em OC-Greedy} against the following algorithms:
\begin{itemize}
    \item \textbf{\rkc-Gonzalez:} A fast heuristic that modifies Gonzalez's algorithm by selecting the $(z+1)^{\text{th}}$ farthest point instead of the farthest at each iteration until we pick $k$ centers. After choosing $k$ centers, it assigns each point to its nearest center and removes the $z$ points with the highest assignment costs.
    
    \item \textbf{Charikar et al.~\cite{charikar2001}:} The $3$-approximation combinatorial algorithm, reported at its achieved solution radius.
    
    \item \textbf{Ding et al.~\cite{ding2019}:} The state-of-the-art randomized algorithm. In their original empirical study, while the authors evaluated the final clustering radius by discarding exactly $z$ outliers, they relaxed their candidate pool during the algorithm's execution to sample from the $2z$ farthest points (setting $\epsilon=1$) to maintain a $1/2$ probability of success per step, requiring $\mathcal{O}(2^k)$ trials. To ensure a fair comparison against algorithms that strictly respect the outlier budget $z$ during selection, we evaluate the tightest formulation of their candidate selection by setting their distance tolerance to $\epsilon=1/z$. Under this strict parameterization, formally guaranteeing $k-1$ successful probabilistic choices theoretically requires an intractable $T = \lceil \ln(100)(\frac{N}{N-z}) (z+1)^{k-1} \rceil$ independent trials. To adapt this for a fair practical evaluation, we cap the algorithm at an empirical budget of $T=\lceil(z{+}1)\ln 100\rceil$ independent trials per configuration. This specific budget formally guarantees a 99\% probability of success for a single probabilistic step, acting as a scalable, linearly growing heuristic cap that allows the baseline to run in practical time frames (executing up to 18,423 independent trials for $z/N=20\%$ at $N=20{,}000$). Like the original authors, we evaluate the final radius of the returned centers by strictly discarding exactly $z$ outliers.    
\end{itemize}

\subsection{Results}

\begin{table*}[htbp!]
\caption{Aggregate LP approximation ratios and runtimes over 36 UCI Adult configurations. Ratio is defined as (Achieved Radius / LP Radius). Lower is better.}
\label{tab:adult}
\centering
\begin{tabular}{lcccccc}
\toprule
Algorithm & Mean Ratio & Std. Dev. & Min Ratio & Max Ratio & Med. Time (s) & Max Time (s) \\
\midrule
LP Oracle & 1.000 & - & 1.000 & 1.000 & 4{,}586 & 338{,}755 \\
\rkc-Gonzalez & 1.280 & 0.049 & 1.197 & 1.370 & 0.002 & 0.02 \\
Ding & 1.235 & 0.032 & 1.179 & 1.331 & 2.22 & 431 \\
Charikar & 1.173 & 0.056 & 1.104 & 1.306 & 3.60 & 564 \\
OC-Greedy & 1.158 & 0.047 & 1.104 & 1.283 & 0.26 & 52 \\
\textbf{OverclusterDRG} & \textbf{1.147} & 0.047 & 1.088 & 1.265 & \textbf{0.36} & \textbf{57} \\
\bottomrule
\end{tabular}
\end{table*}

\paragraph{\textbf{Clustering Quality.}}

\begin{figure}[htbp!]
\centering
  \includegraphics[width=0.8\linewidth]{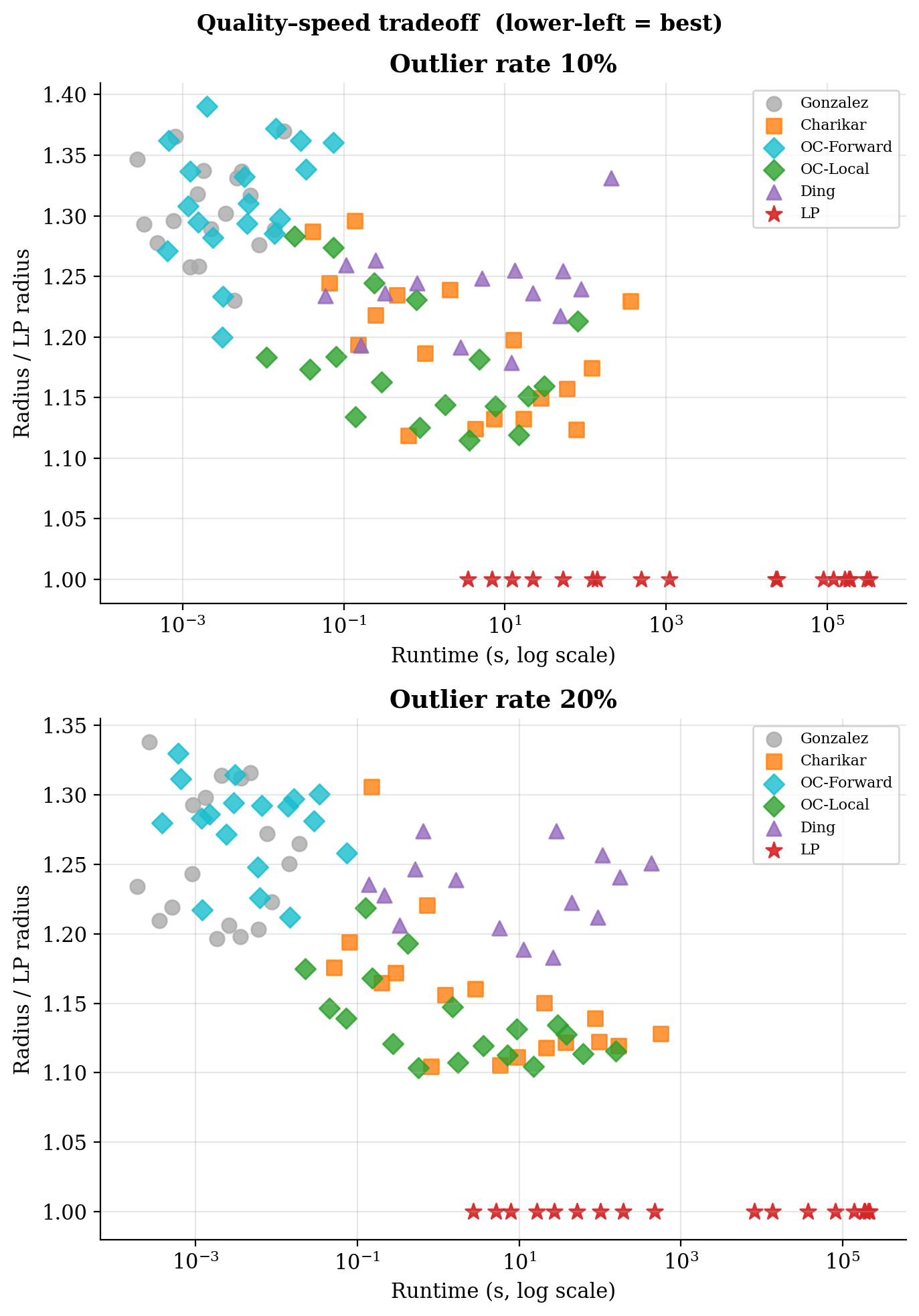} 
  \caption{Quality-runtime tradeoff on UCI Adult. Runtime is on a log axis. OverclusterDRG (green) strictly occupies the lower-left frontier.}
  \label{fig:qualityspeed}
\end{figure}

\begin{figure}[htbp!]
  \centering
  \includegraphics[width=0.8\linewidth]{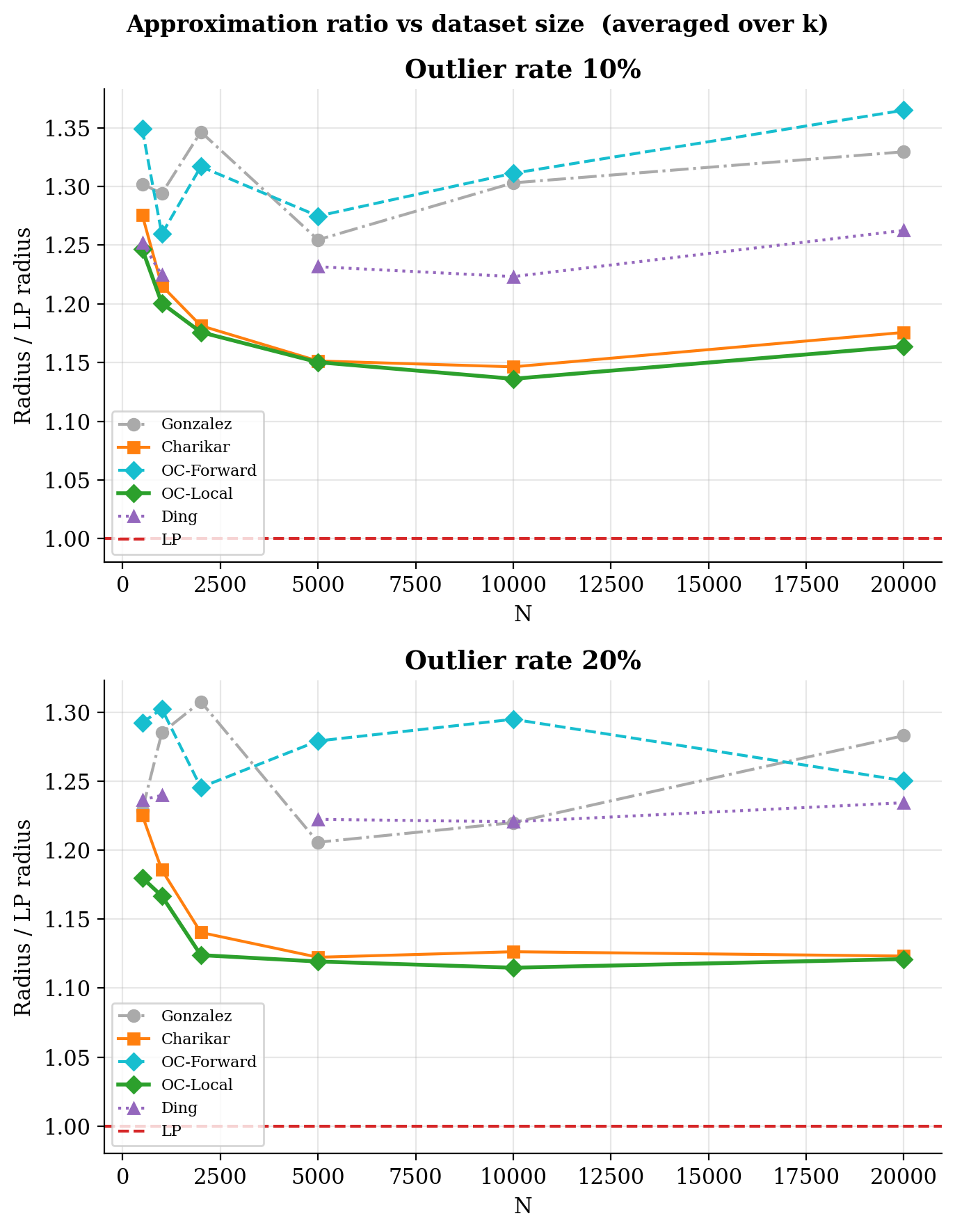} 
  \caption{LP ratio vs dataset size $N$. OverclusterDRG tightly tracks the LP bound, widening the gap against baselines as $N$ scales.}
  \label{fig:ratiovsN}
\end{figure}

The fundamental tension in robust clustering is perfectly captured by our baselines: in the Adult dataset, Charikar achieves an 8.32\% tighter radius than \rkc-Gonzalez on average, but at $5{,}528\times$ the runtime cost, rendering it computationally infeasible beyond $N=5{,}000$. OverclusterDRG successfully bridges this gap. Table~\ref{tab:adult} summarizes the results on the Adult dataset. OverclusterDRG attains the tightest mean approximation ratio ($1.147$), outperforming Charikar ($1.173$), Ding ($1.235$), and \rkc-Gonzalez ($1.280$). On a per-configuration basis, OverclusterDRG beats Charikar in 33/36 cases and beats Ding in 35/36 cases. Figure~\ref{fig:qualityspeed} shows that OverclusterDRG sits firmly on the optimal lower-left frontier of the quality-speed tradeoff. Figure~\ref{fig:ratiovsN} demonstrates that this advantage is highly stable: OverclusterDRG tracks the LP bound closest at every scale up to $N=20{,}000$, while the gap to baselines widens.

\begin{table*}[t]
\caption{LP approximation ratios on secondary datasets. Diabetes ($d{=}8$, 36 configurations), Covertype ($d{=}10$, 30 configurations).}
\label{tab:multidata}
\centering
\begin{tabular}{lcccc c lcccc}
\toprule
\multicolumn{5}{c}{\textbf{Diabetes}} & & \multicolumn{5}{c}{\textbf{Covertype}} \\
\cmidrule(r){1-5} \cmidrule(l){7-11}
Algorithm & Mean & Std & Min & Max & & Algorithm & Mean & Std & Min & Max \\
\midrule
\rkc-Gonzalez & 1.923 & 0.256 & 1.449 & 2.414 & & \rkc-Gonzalez & 1.671 & 0.173 & 1.418 & 1.960 \\
Ding & 1.227 & 0.048 & 1.143 & 1.330 & & Ding & 1.213 & 0.049 & 1.121 & 1.300 \\
Charikar & 1.201 & 0.044 & 1.145 & 1.313 & & Charikar & 1.227 & 0.060 & 1.134 & 1.374 \\
\textbf{OverclusterDRG} & \textbf{1.155} & 0.043 & 1.093 & 1.275 & & \textbf{OverclusterDRG} & \textbf{1.171} & 0.051 & 1.095 & 1.294 \\
\bottomrule
\end{tabular}
\end{table*}

\begin{figure}[htbp!]
\centering
\includegraphics[width=0.8\linewidth]{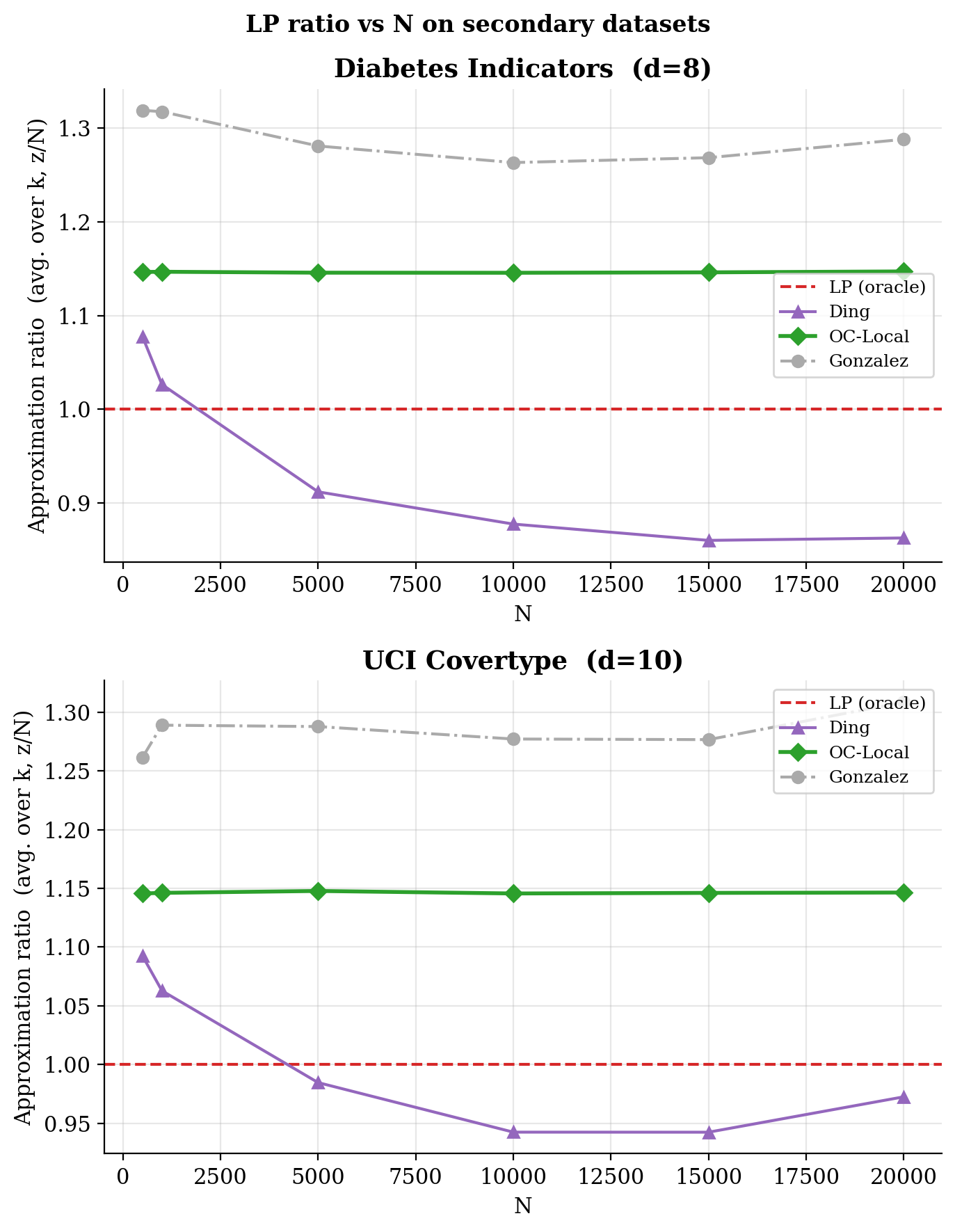}
\caption{LP ratio versus $N$ on Diabetes (left) and Covertype (right). OverclusterDRG tracks the LP bound closely on both datasets.}
\label{fig:ratioboth}
\end{figure}

This dominance consistently replicates across the secondary datasets (Table~\ref{tab:multidata} and Figure~\ref{fig:ratioboth}). OverclusterDRG attains the best mean LP ratio on both Diabetes ($1.155$) and Covertype ($1.171$). The naive \rkc-Gonzalez baseline degrades sharply on these datasets (ratios up to $1.96$).

Paired $t$-tests and Wilcoxon signed-rank tests (Table~\ref{tab:ttests}) confirm that quality gaps are statistically significant ($p < 10^{-5}$). Crucially, OverclusterDRG outperforms its own OC-Greedy initialization in $36/36$ cases ($p < 10^{-8}$). The 1-swap local search provides continuous quality polish for a trivial median cost of 0.10 seconds, proving its practical necessity.

\begin{table}[ht]
\caption{Paired $t$-tests on per-configuration ratio differences ($n=36$). Positive $\bar d$ means OverclusterDRG achieves a tighter (better) radius.}
\label{tab:ttests}
\centering
\resizebox{\columnwidth}{!}{%
\begin{tabular}{lccc}
\toprule
Comparison & Mean Diff ($\bar d$) & $t$-statistic & $p$-value \\
\midrule
OverclusterDRG vs \rkc-Gonzalez & $+0.133$ & $+18.4$ & $<10^{-18}$ \\
OverclusterDRG vs Ding & $+0.088$ & $+12.3$ & $<10^{-12}$ \\
OverclusterDRG vs Charikar & $+0.027$ & $+5.7$ & $<10^{-4}$ \\
OverclusterDRG vs OC-Greedy & $+0.011$ & $+9.1$ & $<10^{-9}$ \\
\bottomrule
\end{tabular}%
}
\end{table}

\paragraph{\textbf{Scalability.}}
To precisely quantify scalability, we performed a log-log regression of runtime against $N$ for all algorithms (Table~\ref{tab:scaling}). OverclusterDRG achieves an empirical scaling of $\mathcal{O}(N^{1.85})$, successfully bridging the speed gap between Gonzalez ($\mathcal{O}(N^{0.80})$) and Charikar ($\mathcal{O}(N^{2.12})$).

\begin{table}[ht]
\caption{Empirical runtime scaling from log-log regression ($T \propto N^b$).}
\label{tab:scaling}
\centering
\resizebox{0.8\columnwidth}{!}{%
\begin{tabular}{lcc}
\toprule
Algorithm & Exponent $b$ & $R^2$ \\
\midrule
\rkc-Gonzalez & 0.80 & 0.939 \\
Ding (Practical budget) & 1.10 & 0.962 \\
\textbf{OverclusterDRG} & \textbf{1.85} & \textbf{0.976} \\
Charikar & 2.12 & 0.999 \\
LP Oracle & 3.40 & 0.988 \\
\bottomrule
\end{tabular}%
}
\end{table}

\begin{table}[ht]
\caption{OverclusterDRG speedup over Ding's heuristic trial count. OverclusterDRG is faster in all 66 configurations.}
\label{tab:speedup}
\centering
\resizebox{\columnwidth}{!}{
\begin{tabular}{lcccc}
\toprule
Dataset & Mean Speedup & Median & Min & Max \\
\midrule
Diabetes ($n{=}36$) & $2.0\times$ & $2.0\times$ & $1.3\times$ & $2.4\times$ \\
Covertype ($n{=}30$) & $2.6\times$ & $2.6\times$ & $2.1\times$ & $3.6\times$ \\
\bottomrule
\end{tabular}
}
\end{table}

Table~\ref{tab:speedup} highlights that OverclusterDRG remains $2\times$ to $3\times$ faster than Ding's truncated heuristic on the secondary datasets. At the hardest Adult configuration tested ($N=20{,}000$, $k=50$, $z=20\%$), Ding's formal theory requires $\sim 10^{15}$ trials. Even under our truncated empirical budget, Ding's heuristic still required $18{,}423$ trials taking $431$ seconds. In contrast, OverclusterDRG finishes its strictly deterministic pass in just $57$ seconds ($7.5\times$ speedup) while attaining a tighter solution quality. 

At extreme scales (Table~\ref{tab:scale}), where the LP and Charikar are intractable, OverclusterDRG remains highly efficient. At $N=99{,}000$, OverclusterDRG finishes in $1{,}100$ seconds, compared to $3{,}800$ seconds for Ding.

\begin{table}[ht]
\caption{Runtimes at large $N$ on Diabetes ($k=20$, $z=10\%$).}
\label{tab:scale}
\centering
\resizebox{\columnwidth}{!}{
\begin{tabular}{rccc}
\toprule
$N$ & OverclusterDRG radius & OverclusterDRG time & Ding time \\
\midrule
20{,}000 & 2.060 & 57s & 133s \\
50{,}000 & 2.110 & 310s & 900s \\
99{,}000 & 2.150 & 1{,}100s & 3{,}800s \\
\bottomrule
\end{tabular}
}
\end{table}

\begin{figure}[ht]
\centering
\includegraphics[width=\linewidth]{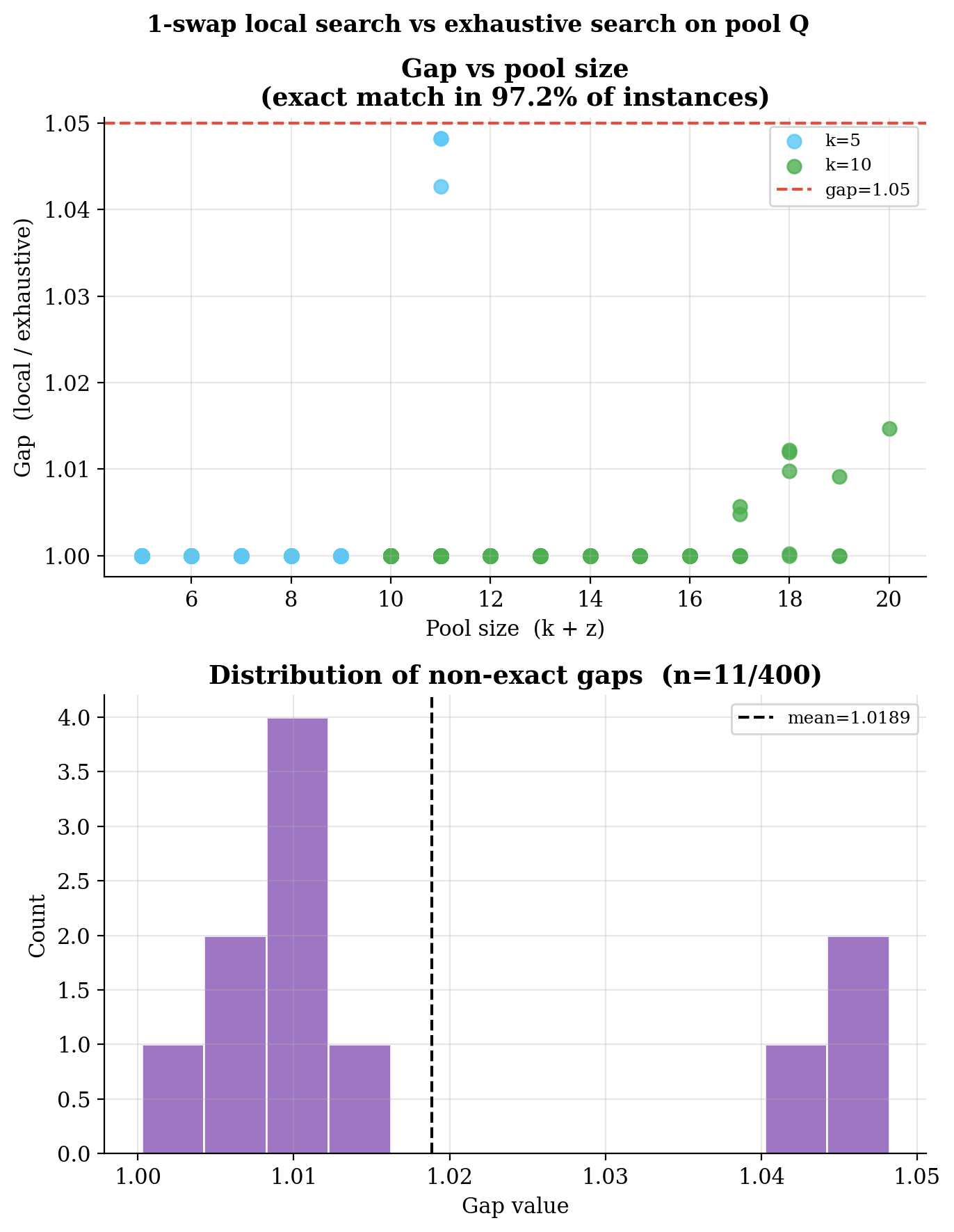}
\caption{Gap between $1$-swap local search and exhaustive pool search across 400 small instances. Local search is exact in $97.2\%$ of instances and never exceeds the exhaustive optimum by more than $4.82\%$.}
\label{fig:localexhaustive}
\end{figure}

\paragraph{\textbf{Validating Local Search Mechanics.}} To quantify the susceptibility of our 1-swap local search to suboptimal local minima, we compared it against an exhaustive $\binom{k+z}{k}$ combinatorial subset evaluation over the pool. On 400 random subsamples of UCI Adult ($N \le 1000$, $k \in \{5, 10\}$, $z \le 10$), OverclusterDRG recovered the exact combinatorial pool optimum in 389 instances (97.2\%). In the 11 instances where the local search did terminate at a suboptimal local minimum, the maximum observed optimality gap was merely 4.82\% (Figure~\ref{fig:localexhaustive}). Crucially, exhaustive combinatorial evaluation required up to 10 seconds on the worst-case instances, whereas OverclusterDRG finished the same extraction in under 3 milliseconds. Thus, the empirical evidence demonstrates that while 1-swap local optima exist, they are exceptionally rare in practical data and their quality penalty is tightly bounded, acting as a efficient candidate extraction mechanism.

\subsection{Choice of \texorpdfstring{$z$}{z} for Overpicking Centers}
\label{sec:oablation}

Our algorithm fundamentally relies on provisioning the initial candidate pool with extra over picking steps. We swept the number of over picked centers across 72 configurations (on Diabetes and Covertype) to validate our choice of over picking by exactly $z$ candidates.

\begin{figure}[htbp!]
\centering
\includegraphics[width=\linewidth]{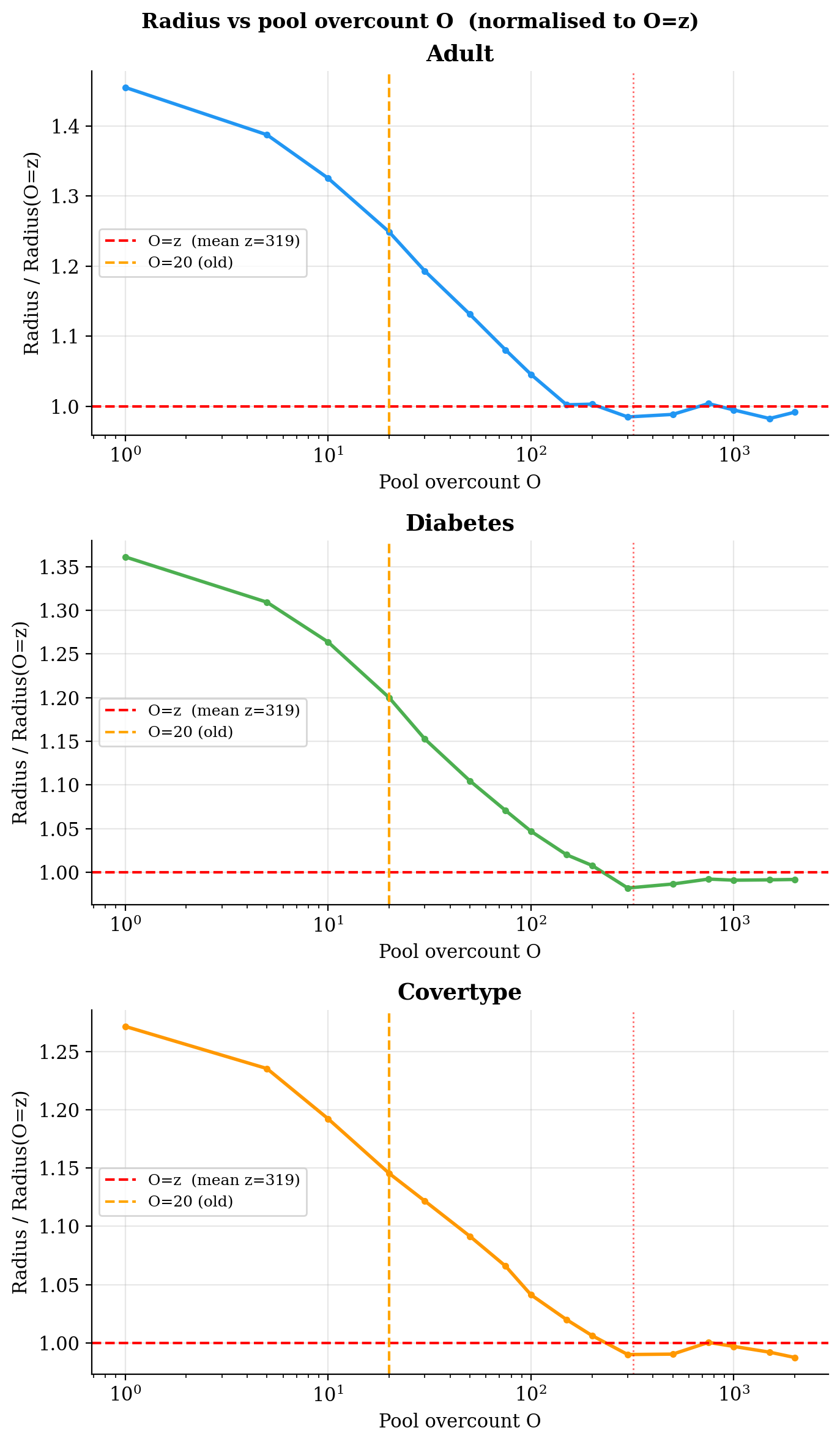}
\caption{Radius (normalized) as a function of the number of over picked centers on a log axis. Quality degrades sharply when over picking $< z$, and flattens once it reaches $z$.}
\label{fig:oablation}
\end{figure}

Figure~\ref{fig:oablation} reveals a sharp behavioral phase transition. The resulting radius degrades severely when the pool is starved (over picking $< z$) because unhandled outliers successfully hijack the Gonzalez traversal. However, exactly as the number of over picked centers reaches $z$, the quality flattens to a tight plateau. 

\begin{table}[htbp!]
\caption{Impact of the over picking parameter on clustering quality (measured across 72 configurations).}
\label{tab:osettings}
\centering
\resizebox{\columnwidth}{!}{
\begin{tabular}{lccc}
\toprule
Setting & Within 5\% & Mean gap & Max gap \\
\midrule
Over picking by $z$ & 53/72 & $+3.7\%$ & $+11.8\%$ \\
Over picking by $2z$ & 69/72 & $+1.3\%$ & $+8.2\%$ \\
\bottomrule
\end{tabular}
}
\end{table}

Table~\ref{tab:osettings} quantifies this: over picking by exactly $z$ centers lands within $5\%$ of the absolute best achievable quality in 53 of 72 configurations, with a mean gap of just $3.7\%$. Pushing the overcount to $2z$ marginally increases quality, but doubles the size of the precomputed distance matrix and proportionally slows down the local search. 

\begin{figure}[htbp!]
\centering
\includegraphics[width=\linewidth]{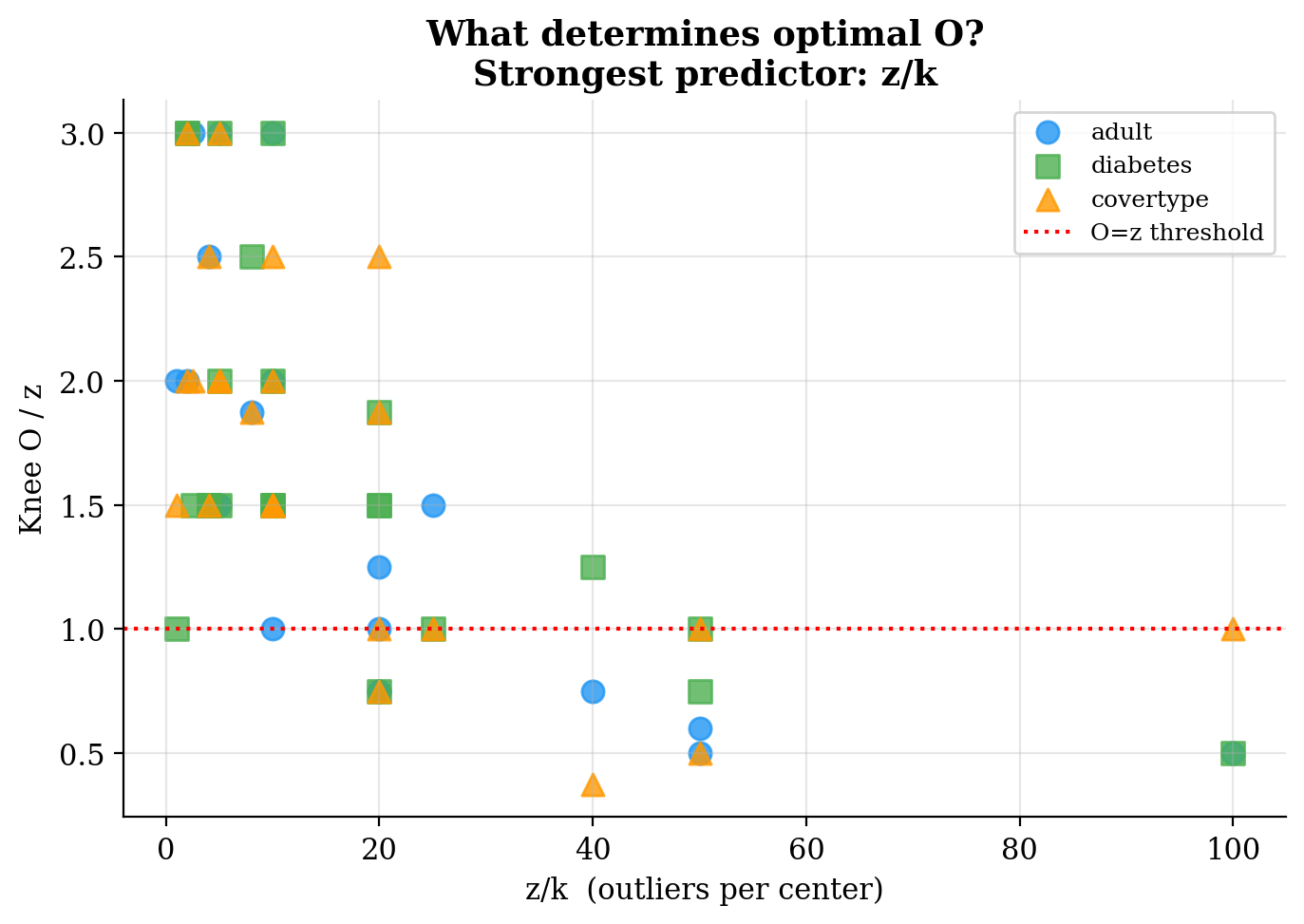}
\caption{The empirical knee plotted against $z/k$. More outliers per center push the knee tightly toward $z$ itself.}
\label{fig:ovszk}
\end{figure}

A correlation analysis against the precise location of the "knee" (where improvements flatten below 2\%) found that the ratio $z/k$ is the strongest structural predictor (Pearson $r=-0.60, p < 10^{-9}$). When each true cluster center is responsible for many outliers, the knee tightly hugs $z$ (Figure~\ref{fig:ovszk}).

\begin{figure*}[htbp!]
\centering
\includegraphics[width=\linewidth]{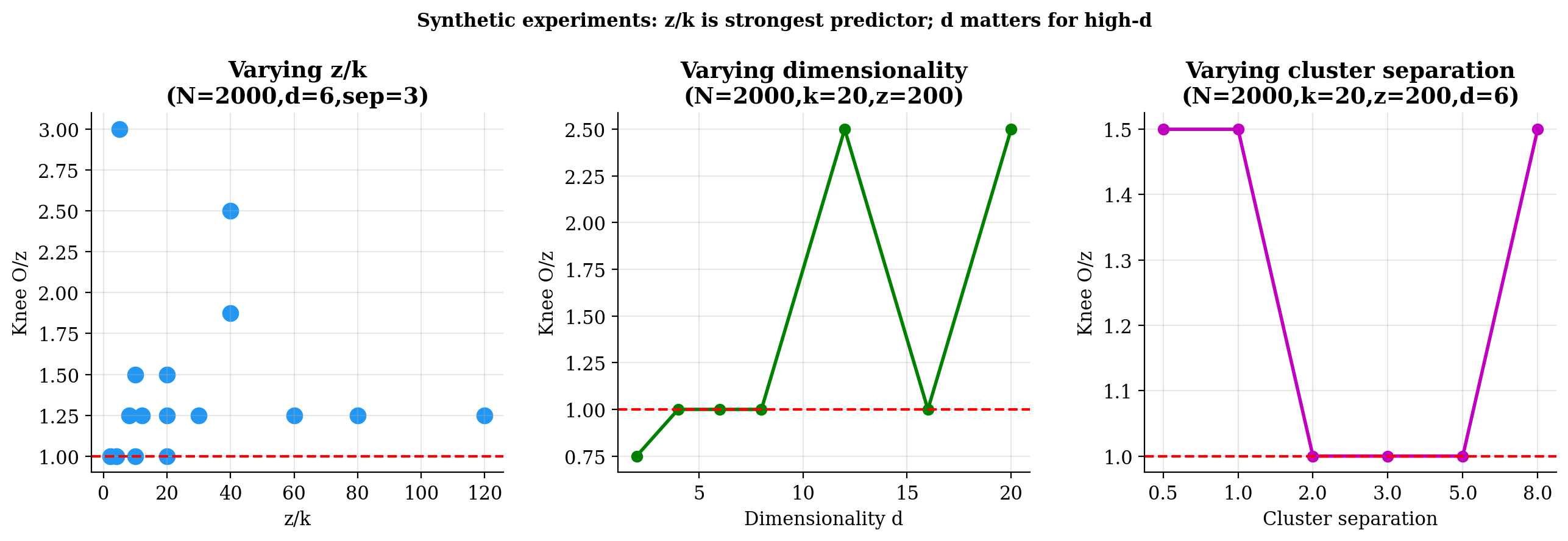}
\caption{Controlled swaps on synthetic data. Varying the $z/k$ ratio controls the transition point (left). Dimensionality (center) and cluster separation (right) have little to no effect, keeping knee strictly bounded near $z$.}
\label{fig:synthetic}
\end{figure*}

To confirm this structural dependence without real-world confounding factors, we conducted controlled synthetic swaps (Figure~\ref{fig:synthetic}). Across two decades of varying $z/k$ (left), varying cluster separation from tight to well-separated (right), and sweeping dimensionality from 2 to 20 (center), the knee never strays far from $z$. Only extreme high dimensionality ($d \ge 12$) pushes the plateau slightly rightward toward $2.5z$. Consequently, over picking by exactly $z$ centers acts as the optimal structural and computational threshold.

\section{Conclusion}
\label{sec:conclusion}

OverclusterDRG fills the gap between quality-focused and speed-focused algorithms for the Robust $k$-Center problem. By decoupling the combinatorial outlier-selection task from the clustering task through deterministic farthest-point oversampling, we created a heuristic that is scalable and LP-free. Our empirical evaluations demonstrate that OverclusterDRG attains a mean LP ratio of 1.147 on the UCI Adult dataset and 1.15 - 1.17 on two secondary datasets, consistently beating the solution qualities of exact combinatorial approximation schemes (Charikar et al.) and randomized methods (Ding et al.). Furthermore, it achieves these tighter bounds at a $7\times$ to $8\times$ median speedup, remaining deterministic and highly efficient even at $n \approx 10^5$, where LP baselines are entirely unavailable.

\newpage
\bibliographystyle{siamplain}
\bibliography{refs}
\end{document}